\documentclass{article}

\usepackage[final]{neurips_2019}
\usepackage{graphicx}

\usepackage[utf8]{inputenc} 
\usepackage[T1]{fontenc}    
\usepackage{hyperref}       
\usepackage{url}            
\usepackage{booktabs}       
\usepackage{amsfonts}       
\usepackage{nicefrac}       
\usepackage{microtype}      
\usepackage{floatrow}
\usepackage[T1]{fontenc}
\usepackage[font=small,labelfont=bf,tableposition=top]{caption}
\usepackage{diagbox}
\usepackage[para]{footmisc}

\title{\vspace{-2mm}Domain-Relevant Embeddings \\ for Medical Question Similarity\vspace{-2mm}}

\author{
    Clara H.~McCreery \\ 
    \url{mccreery@cs.stanford.edu}  
    \And
    Namit Katariya \\
    \url{namit@curai.com}
    \And
    Anitha Kannan \\
    \url{anitha@curai.com}
    \And
    Manish Chablani \\
    \url{manish@curai.com}
    \And
    Xavier Amatriain \\
    \url{xavier@curai.com}
}

\begin{document}

\maketitle
\vspace{-3mm}
\begin{abstract}
The rate at which medical questions are asked online far exceeds the capacity of qualified people to answer them, and many of these questions are not unique. Identifying same-question pairs could enable questions to be answered more effectively. While many research efforts have focused on the problem of general question similarity for non-medical applications, these approaches do not generalize well to the medical domain, where medical expertise is often required to determine semantic similarity. In this paper, we show how a semi-supervised approach of pre-training a neural network on \emph{medical question-answer pairs} is a particularly useful intermediate task for the ultimate goal of determining medical question similarity. While other pre-training tasks yield an accuracy below 78.7\% on this task, our model achieves an accuracy of 82.6\% with the same number of training examples, and an accuracy of 80.0\% with a much smaller training set.

\end{abstract}

\vspace{-2mm}
\section{Introduction}

With the ubiquity of the Internet and the emergence of medical question-answering websites such as ADAM \footnote{\url{www.adam.com}}, WebMD \footnote{\url{www.webmd.com}}, and HealthTap \footnote{\url{www.healthtap.com}}, people are increasingly searching online for answers to their medical questions. However, the number of people asking medical questions online far exceeds the number of qualified medical experts -- i.e doctors -- answering them, leaving many questions inadequately answered. One way to address this problem is to build a system that can automatically match \emph{unanswered} questions to \emph{answered} questions with the same meaning, or mark them as priority for a doctor if no similar answered questions exist. This approach uses doctor time more efficiently, reducing the number of unanswered questions and lowering the cost of providing care online. 

Coming up with an accurate algorithm for finding similar medical questions, however, is difficult. Simple heuristics such as word-overlap are ineffective because \emph{Can a menstrual blood clot travel to your heart or lungs like other blood clots can?} and \emph{Can clots from my period cause a stroke or embolism?} are similar questions with low overlap but \emph{Is candida retested after treatment} and \emph{Is Chlamydia retested after treatment?} are critically different and only one word apart. In this paper we address the question of how we can modify the latest general NLP models to work better for this task.

Given the recent success of pre-trained bi-directional transformer networks for NLP outside the medical field \citep{Peters_2018, devlin2018bert, radford2019language, yang2019xlnet}, most research efforts in medical NLP have tried to apply these general language models to medical tasks \citep{ben-abacha-etal-2019-overview}. However, these models are not trained on medical information, and make errors that reflect this. In this work, we augment the features in these general language models using the depth of information that is stored within a medical question-answer pair to provide the needed medical knowledge. We show how pre-training a transformer network on a medical question-answer matching task embeds relevant medical knowledge into the model that it otherwise does not have.

The main contributions of this paper are:
\begin{itemize}
  \item We prove that, particularly for medical NLP, domain matters: pre-training on a different task in the same domain outperforms pre-training on the same task in a different domain
  \item We show that the difference in performance between in-domain and out-of domain models is exacerbated when the final task has a small train set
  \item We analyze the examples that the out-of-domain models label incorrectly and identify that most errors are in fact due to unknown medical terms. In contrast, the in-domain models are able to label many of these examples correctly.
  
\end{itemize}
\vspace{-2mm}
\section{Data}

\subsection{Medical Question Pairs}
\label{data:qq}

Our doctors hand-generate a small dataset of medical question pairs (3,000 pairs total). We explicitly choose doctors for this task because determining whether or not two medical questions are the same requires medical training that crowd-sourced workers rarely have. We present doctors with a list of patient-asked questions from HealthTap \citep{durakkerem}, and for each provided question, ask them to:
\begin{enumerate}
    \item Rewrite the original question in a different way while maintaining the same intent. Restructure the syntax as much as possible and change medical details that would not impact your response (ex.`I'm a 22-y-o female' could become `My 26 year old daughter' ).
    \item Come up with a related but dissimilar question for which the answer to the original question would be WRONG OR IRRELEVANT. Use similar key words. 
\end{enumerate}

The first instruction generates a \textit{positive} question pair (match) and the second generates a \textit{negative} question pair (mismatch). With the above instructions, we intentionally frame the task such that positive question pairs can look very different by superficial metrics, and negative question pairs can conversely look very similar. This ensures that the task is not artificially easy.

We anticipate that each person interprets these instructions slightly differently, so no doctor providing data in the train set is used to generate any data in the test set. This should reduce bias. To obtain an oracle score, we have doctors hand-label question pairs that a different doctor generated. The accuracy of the second doctor with respect to the labels intended by the first is used as an oracle and is 87.6\% in our test set of 836 question pairs.

\subsection{Pre-Training Tasks}

\textbf{Question Answer Pairs (QA)}
The training data for medical question-answer pairs comes from HealthTap as well. However, all of the questions that are used in the question pairs dataset are removed from the training data for the intermediate task. We isolate each true question-answer pair from the HealthTap data and label these as positive examples. We then keep each question and pair it with a random answer from the same main category and label these as negative examples. 

\textbf{Quora Question Pairs (QQP)} Our out-of-domain question pairs come from the general question-answer forum, Quora \citep{csernai_2017}.
\vspace{-2mm}
\section{Problem Setup}

\begin{figure}[ht]
\begin{center}
\includegraphics[width=\textwidth]{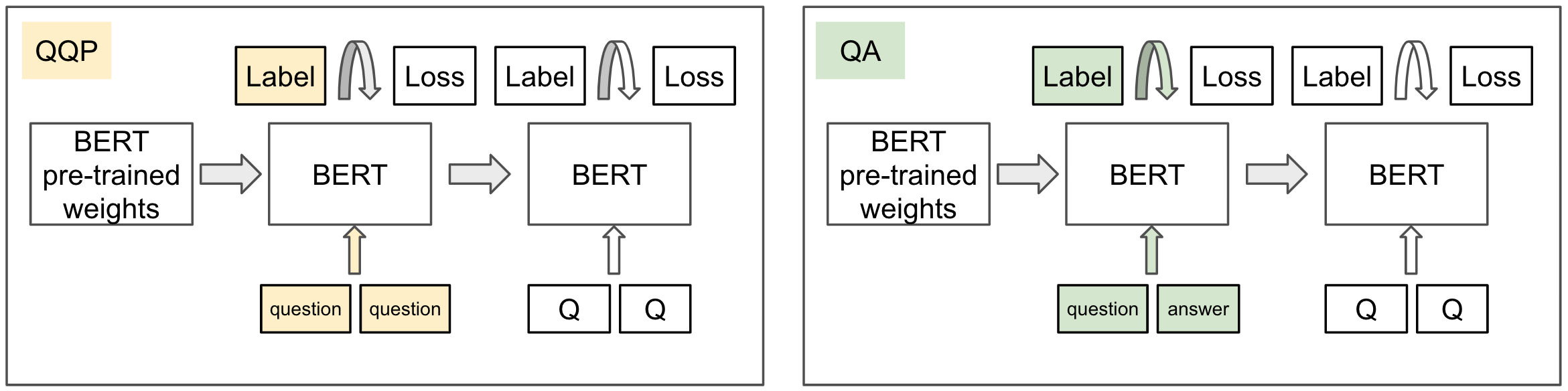}
\end{center}
\caption{We perform a double finetune from BERT to an intermediate task to our medical question-similarity task for two different intermediate tasks: quora question-question pairs (left) and medical question-answer pairs (right)}
\label{fig:process}
\end{figure}

We seek to understand how to best transfer relevant knowledge to a general language model for medical question similarity. Our hypothesis is that by training on a large corpus for a similar medical task, we can embed medical knowledge into the model. Our approach uses transfer learning from a bi-directional transformer network to get the most out of a small medical question pairs dataset. 

We start with the architecture and weights from BERT \citep{devlin2018bert} and finetune twice; first on an intermediate task and then on the final task of medical question pairs. We do this for two different intermediate tasks: quora question similarity (QQP) and medical question answer pairs (QA) (Figure~\ref{fig:process}). For a baseline, we skip the intermediate finetune and directly train BERT on our small medical question-pairs dataset. To verify our results, we repeat this experiment with the XLNet architecture \citep{yang2019xlnet}. 

For each intermediate task, we use 363,000 training examples to ensure that differences in performance are not due to dataset sizes. We then finetune each intermediate-task-model on the medical-question pairs until convergence. All experiments are done with 5 different random train/validation splits to generate error bars representing one standard deviation. A learning rate of 2e-5, batch size of 16, and maximum sentence length of 200 tokens is used to train on two parallel NVIDIA Tesla V100 GPUs. Since we generate a perfectly balanced test dataset, we use accuracy as our quantitative metric for comparison.

To get a better qualitative understanding of performance, we perform error analysis. We define a consistent error as one that is made by at least 4 of the 5 models trained on different train/validation splits. Similarly, we consider a model as getting an example consistently correct if it does so on at least 4 of the 5 models trained on different train/validation splits. By investigating the question pairs that a model-type gets consistently wrong, we can form hypotheses about why the model may have gotten that specific question pair wrong. Then, by making small changes to the input until the models label those examples correctly, we can validate or disprove these hypotheses.
\vspace{-2mm}
\section{Results}
\label{results-section}

\begin{minipage}{\textwidth}
\begin{minipage}[b]{0.55\textwidth}
\centering
\captionsetup{type=figure}
\includegraphics[width=1.1\textwidth]{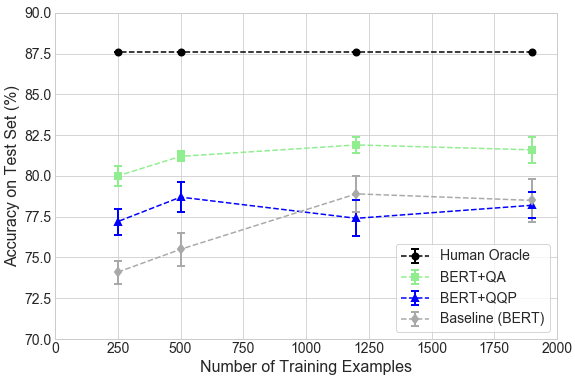}
\captionof{figure}{The intermediate task of training on question-answer pairs (BERT+QA) reliably outperforms training on Quora question pairs (BERT+QQP). Differences are exacerbated with fewer training examples in the final task.}
\end{minipage}
\hfill
\begin{minipage}[b]{0.4\textwidth}
\centering
    \small\addtolength{\tabcolsep}{-1pt}
    \captionsetup{type=table}
    \begin{tabular}{lll}
    \multicolumn{1}{c}{\bf Model} &\multicolumn{1}{c}{\bf BERT} &\multicolumn{1}{c}{\bf XLNet} \\ 
    \multicolumn{1}{c}{\bf Num Train} &\multicolumn{1}{c}{\bf 1.9k} &\multicolumn{1}{c}{\bf 1.9k} \\
    \hline \\
    \multicolumn{1}{l}{\bf Intermediate Task} & & \\
    None & 78.5\% & 77.7\% \\
    & $\pm$ 2.1\%  & $\pm$ 1.3\% \\ \\
    Quora (QQP) & 78.2\% & 78.2\% \\
    & $\pm$ 0.2\% & $\pm$ 0.8\% \\ \\
    Medical QA & \textbf{81.6\%} & \textbf{82.6\%} \\
    & $\pm$ 0.8\% & $\pm$ 0.8\% \\ \\
    \\
    \end{tabular}
  \captionof{table}{The same trends hold across different choices of model architecture (BERT and XLNet)}
  \label{table:results}
\end{minipage}
\end{minipage}

\subsection{Domain Matters}

The question we are trying to answer is whether domain of the training corpus matters more than task-similarity when choosing an intermediate training step for the medical question similarity task. Accuracy on the final task (medical question similarity) is our quantitative metric for comparing each of the models.

We finetune BERT on the intermediate tasks of Quora question similarity (QQP) and medical question-answer pairs (QA) first, to compare the relative importance of task-similarity and domain-similarity. For each model, we then finetune on the final task. We find that across multiple final-task training set sizes the in-domain question-answer model reaches a higher question-similarity accuracy than the QQP model does, with an average difference in performance of 3.7\% (Table~\ref{table:results}). Furthermore, the smaller the training set size for the final task, the more exacerbated these differences become.

Qualitatively, we see that the mistakes made consistently by the baseline model and the QQP-trained model often require some medical knowledge, and the QA-trained model gets some of these examples right. So, both quantitatively and qualitatively, we see that models trained on a related in-domain task (medical question-answering) outperforms models trained on the same question-similarity task but an out-of-domain corpus (quora question pairs).

\subsection{Error Analysis}

\begin{table}[ht]
\caption{Example error analysis in which an example is modified until each model labels it consistently correct}
\label{fig:errors}
\begin{center}
\includegraphics[width=\textwidth]{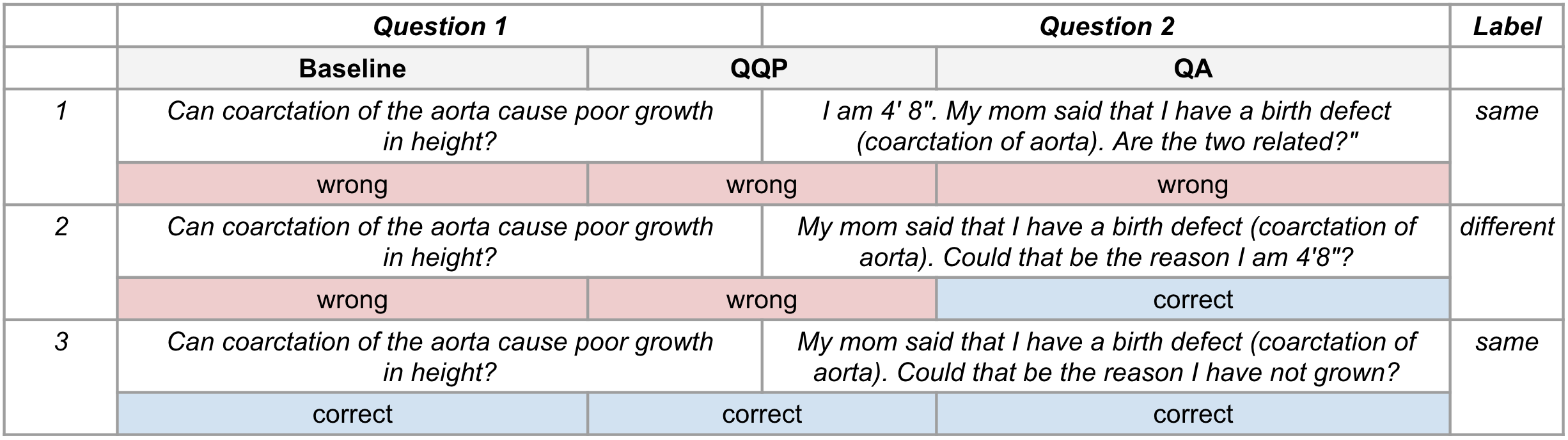}
\end{center}
\end{table}

From looking at the question pairs that our models get wrong, we can form hypotheses about why each example is mislabeled. By augmenting them to remove one difficult aspect of the question pair at a time, we can then prove or disprove these hypotheses. 

Consider the example in Table \ref{fig:errors}. In order to label this example correctly as it is written in row 1, the model has to understand the syntax of the question and know that 4'8" in this context represents poor growth. Changing the second question to what is written in row 2 prompts the QA model to label it correctly, indicating that what the QA model was misunderstanding was the question's phrasing. In contrast, it is not until we replace \textit{I am 4'8"} with \textit{I have not grown} as shown in row 3 that the out-of-domain models get this example correct.  So, while numerical reasoning was the difficult part of that question pair for the baseline and QQP models, the question answer model was actually able to understand that 4'8" is a short height. This supports the claim that pre-training on the medical task of question answering embedded medically relevant information into the model that the out-of-domain models are still missing. More examples of errors we analyzed can be found in Appendix~\ref{appendixA}.

\vspace{-2mm}
\section{Conclusions}
\label{conclusions}

In this work, we show the importance of domain: pre-training on a different task in the same domain outperforms pre-training on the same task from a different domain. We also observe that this difference is amplified for smaller final-task training sets. Through our error analysis, we gain insight into the types of mistakes that our best models make. This will enable the use of weak supervision and augmentation rules to supplement our training data with more examples in those error-prone regions. Our error analysis further shows that the QQP model learns general question synonyms and equivalent phrasings in questions, but it is more prone to making mistakes due to medical synonyms. Although the QA model outperforms the QQP model, there are a few examples where the QQP model seems to have learned information that is missing from the QA model. In the future, we plan to explore whether we can combine the features learned from these types of tasks into one model with multi-task learning to get even better performance.



\nocite{*}

\bibliography{neurips_2019}
\bibliographystyle{neurips_2019}

\pagebreak

\appendix
\section{Appendix}
\label{appendixA}

\begin{table}[ht]
\caption{Examples of the types of question pairs that each model missed- inclusion of a misspelled example is intentional and representative of the real data}
\label{fig:errors_appendix}
\begin{center}
\includegraphics[width=\textwidth]{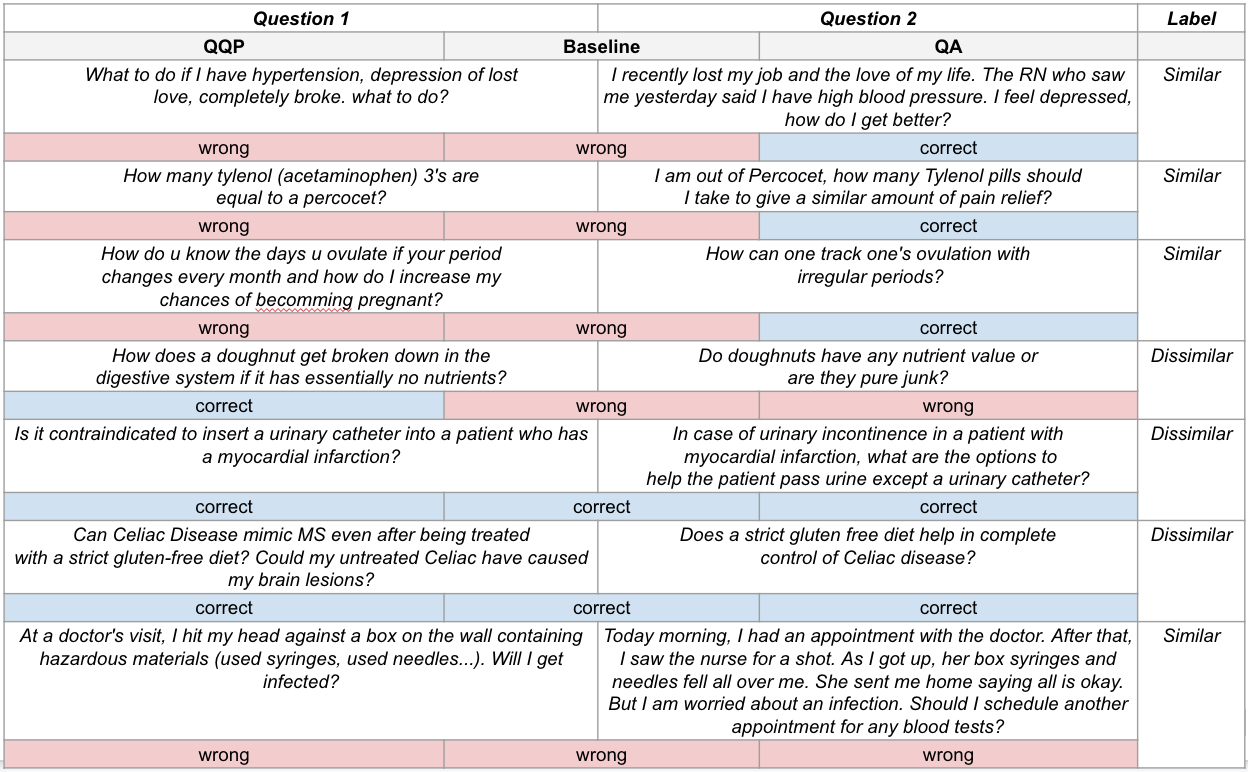}
\end{center}
\end{table}

\end{document}